# A Minimum Relative Entropy Controller for Undiscounted Markov Decision Processes

Pedro A. Ortega                               peortega@dcc.uchile.cl
Daniel A. Braun                               dab54@cam.ac.uk
Dept. of Engineering, University of Cambridge, Cambridge CB2 1PZ, UK

## Abstract

Adaptive control problems are notoriously difficult to solve even in the presence of plant-specific controllers. One way to by-pass the intractable computation of the optimal policy is to restate the adaptive control as the minimization of the relative entropy of a controller that ignores the true plant dynamics from an informed controller. The solution is given by the Bayesian control rule—a set of equations characterizing a stochastic adaptive controller for the class of possible plant dynamics. Here, the Bayesian control rule is applied to derive BCR-MDP, a controller to solve undiscounted Markov decision processes with finite state and action spaces and unknown dynamics. In particular, we derive a non-parametric conjugate prior distribution over the policy space that encapsulates the agent's whole relevant history and we present a Gibbs sampler to draw random policies from this distribution. Preliminary results show that BCR-MDP successfully avoids sub-optimal limit cycles due to its built-in mechanism to balance exploration versus exploitation.

## 1. Introduction

Adaptive control problems, i.e. the design of controllers for plants with unknown dynamics, are notoriously difficult. Even when the plant dynamics is known to belong to a particular class for which optimal controllers are available, constructing the corresponding optimal adaptive controller is in general intractable (Duff, 2002). Thus, virtually all of the effort of the research community is centered around the development of tractable approximations.

Recently, new formulations of the adaptive control problem that are based on the minimization of a relative entropy criterion have attracted the interest of the reinforcement learning (RL) community. For example, it has been shown that a large class of optimal control problems can be solved very efficiently if the problem statement is reformulated as the minimization of the deviation of the dynamics of a controlled system from the uncontrolled system (Todorov, 2006; 2009; Kappen et al., 2009). A similar approach minimizes the deviation of the causal input/output-relationship of a Bayesian mixture of controllers from the true controller, obtaining an explicit solution called the *Bayesian control rule* (Ortega & Braun, 2010). This control rule is particularly interesting because it leads to stochastic controllers that infer the optimal controller on-line by combining the plant-specific controllers, implicitly using the uncertainty of the dynamics to trade-off exploration versus exploitation.

Markov decision processes (MDPs) with undiscounted/averaged rewards constitute an important problem class in RL that has been far less studied than their discounted counterpart. While discounted rewards are suitable in many applications, a wide variety of tasks—such as those found in control tasks where the optimal trajectory is a limit cycle, e.g. network load balancing, automatic assembly, queue management and control of embedded systems—are more naturally stated in terms of optimizing the average reward. However, finding an optimal policy for the average reward function is significantly more difficult than the discounted reward. Unlike the discounted case, in undiscounted MDPs the Bellman optimality equations are strongly coupled and the effective horizon is unbounded. A systematic study in Mahadevan (1996) has shown that exploration plays a crucial role in undiscounted MDP algorithms, as insufficient exploration may lead to the convergence to a sub-optimal limit cycle. Several algorithms have been proposed for undiscounted MDPs, most notably R-learning and its variants (Schwartz, 1993; Singh, 1994), which are inspired by Watkins' Q-learning (Watkins, 1989) and are simple to implement; and $E^3$ (Kearns & Singh, 1998) and R-max (Brafman & Tennenholtz, 2001), which are advanced algorithms that attain near-optimal average reward in polynomial time.

The aim of this paper is to demonstrate how the Bayesian control rule can be used to solve adaptive control problems, illustrating its generality and conceptual simplicity. In particular, undiscounted MDPs with finite state and action space and unknown dynamics. We derive an adaptive controller, which we call BCR-MDP, that employs a conjugate prior distribution over the policy space to concisely encapsulate the agent's history and to infer the optimal policy. Furthermore, we introduce a Gibbs sampler implementing the controller.

## 2. Background

### 2.1. Bayesian control rule

Let $\mathcal{O}$ and $\mathcal{A}$ be two finite sets of symbols, where the former is the set of inputs (observations) and the second the set of outputs (actions). Actions and observations at time $t$ are denoted as $a_t \in \mathcal{A}$ and $o_t \in \mathcal{O}$ respectively, and we use the shorthand $a_{\leq t} := a_1, a_2, \ldots, a_t$ and the like to simplify the notation of strings. We assume that the interaction between the controller and the plant proceeds in cycles $t = 1, 2, \ldots$ where in cycle $t$ the controller issues action $a_t$ and the plant responds with an observation $o_t$. A controller is defined as a probability distribution $P$ over the input/output (I/O) stream, and it is fully characterized by the conditional probabilities

$$P(a_t|a_{<t}, o_{<t}) \quad \text{and} \quad P(o_t|a_{\leq t}, o_{<t})$$

representing the probabilities of emitting action $a_t$ and collecting observation $o_t$ given the respective I/O history. Similarly, a plant is defined as a probability distribution $Q$ characterized by the conditional probabilities

$$Q(o_t|a_{\leq t}, o_{<t})$$

representing the probabilities of emitting observation $o_t$ given the I/O history.

If the plant is known, i.e. if the conditional probabilities $Q(o_t|a_{\leq t}, o_{<t})$ are known, then the designer can build a suitable controller by equating the observation streams as $P(o_t|a_{\leq t}, o_{<t}) = Q(o_t|a_{\leq t}, o_{<t})$ and by defining action probabilities $P(a_t|a_{<t}, o_{<t})$ such that the resulting distribution $P$ maximizes a desired utility criterion. We say that $P$ *is tailored to* $Q$. In many cases the conditional probabilities $P(a_t|a_{<t}, o_{<t})$ will be deterministic, but there are situations (e.g. in repeated games) where the designer might prefer stochastic policies instead.

If the plant is unknown then one faces an adaptive control problem. Assume we know that the plant $Q_\theta$ is going to be drawn randomly from a set $\mathcal{Q} := \{Q_\theta\}_{\theta \in \Theta}$ of possible plants indexed by $\Theta$. Assume further we have available a set of controllers $\mathcal{P} := \{P_\theta\}_{\theta \in \Theta}$, where each $P_\theta$ is tailored to $Q_\theta$. How can we now construct a controller $P$ such that its behavior is as close as possible to the tailored controller $P_\theta$ under any realization of $Q_\theta \in \mathcal{Q}$?

A naïve approach would be to minimize the relative entropy of the controller $P$ with respect to the true controller $P_\theta$, averaged over all possible values of $\theta$. However, this is syntactically incorrect. The important observation made in Ortega & Braun (2010) is that we do not want to minimize the deviation of $P$ from $P_\theta$, but the deviation of the causal I/O dependencies in $P$ from the causal I/O dependencies in $P_\theta$. Intuitively speaking, we do not want to predict actions and observations, but to predict the observations (effect) given actions (causes). More specifically, they propose to minimize a set of (causal) divergences $C$ defined by

$$\begin{aligned} C &:= \limsup_{t \to \infty} \sum_\theta P(\theta) \sum_{\tau=1}^t C_\tau \\ C_\tau &:= \sum_{o_{<\tau}} P_\theta(\hat{a}_{<\tau}, o_{<\tau}) C_\tau(\hat{a}_{<\tau}, o_{<\tau}) \\ C_\tau(h) &:= \sum_{a_\tau} \sum_{o_\tau} P_\theta(a_\tau, o_\tau|h) \log \frac{P_\theta(a_\tau, o_\tau|h)}{P(a_\tau, o_\tau|h)}, \end{aligned} \quad (1)$$

where $P(\theta)$ is the prior probability of $\theta \in \Theta$, $\hat{a}_\tau$ denotes an intervened (not observed) action at time $\tau$, and $\hat{a}_1, \hat{a}_2, \hat{a}_3, \ldots$ is an arbitrary sequence of intervened actions.

In Ortega & Braun (2010), it is shown that the controller $P$ that minimizes $C$ in Equation (1) for any sequence of intervened actions is given by the conditional probabilities

$$\begin{aligned} P(a_t|\hat{a}_{<t}, o_{<t}) &:= \sum_\theta P_\theta(a_t|a_{<t}, o_{<t}) P(\theta|\hat{a}_{<t}, o_{<t}) \\ P(o_t|\hat{a}_{\leq t}, o_{<t}) &:= \sum_\theta P_\theta(o_t|a_{\leq t}, o_{<t}) P(\theta|\hat{a}_{<t}, o_{<t}) \end{aligned} \quad (2)$$

where

$$P(\theta|\hat{a}_{\leq t}, o_{\leq t}) := \frac{P_\theta(o_t|a_{\leq t}, o_{<t})P(\theta|\hat{a}_{<t}, o_{<t})}{\sum_{\theta'} P_{\theta'}(o_t|a_{\leq t}, o_{<t})P(\theta'|\hat{a}_{<t}, o_{<t})}. \quad (3)$$

Equations (2) and (3) constitute the *Bayesian control rule*. This result is obtained by using properties of interventions using causal calculus (Pearl, 2000). It is worth to point out that the resulting controller is fully defined in terms of its constituent controllers in $\mathcal{P}$. It is customary to use the notation

$$P(a_t|\theta, a_{<t}, o_{<t}) := P_\theta(a_t|a_{<t}, o_{<t})$$
$$P(o_t|\theta, a_{\leq t}, o_{<t}) := P_\theta(o_t|a_{\leq t}, o_{<t}),$$

that is, treating the different controllers as hypotheses of a Bayesian model. The resulting control law is in general stochastic. Also, note that by construction, an adaptive code for the I/O stream based on the Bayesian control rule is optimal for the class of plants considered (MacKay, 2003).

### 2.2. MDPs

**Definitions.** An *MDP* is defined as a tuple $(\mathcal{X}, \mathcal{A}, T, r)$: $\mathcal{X}$ is the state space; $\mathcal{A}$ is the action space; $T_a(x; x') = \mathbf{Pr}(x'|a, x)$ is the probability that an action $a \in \mathcal{A}$ taken in state $x \in \mathcal{X}$ will lead to state $x' \in \mathcal{X}$; and $r(x, a) \in \mathcal{R} := \mathbb{R}$ is the immediate reward obtained in state $x \in \mathcal{X}$ and action $a \in \mathcal{A}$. The interaction proceeds in time steps $t = 1, 2, \ldots$ where at time $t$, action $a_t \in \mathcal{A}$ is issued in state $x_{t-1} \in \mathcal{X}$, leading to a reward $r_t = r(x_{t-1}, a_t)$ and a new state $x_t$ that starts the next time step $t+1$. Hence, starting from an initial state $x_0 \in \mathcal{X}$, an I/O sequence has the form

$$x_0 \to a_1 \to (r_1, x_1) \to a_2 \to (r_2, x_2) \to \cdots$$
$$\cdots \to a_{t-1} \to (r_{t-1}, x_{t-1}) \to a_t \to (r_t, x_t) \to \cdots$$

A stationary closed-loop control policy $\pi : \mathcal{X} \to \mathcal{A}$ assigns an action to each state. For MDPs there always exists an optimal stationary deterministic policy and thus one only needs to consider such policies. For undiscounted MDPs, the *goal* is to find a policy that maximizes the time-averaged reward $\frac{1}{t}\sum_{\tau=1}^{t} r_\tau$ as $t \to \infty$.

**Bellman optimality equations.** In undiscounted MDPs the average reward per time step for a fixed policy $\pi$ with initial state $x$ is defined as follows: $\rho^\pi(x) = \lim_{t \to \infty} \mathbf{E}^\pi[\frac{1}{t}\sum_{\tau=0}^{t} r_\tau]$. It can be shown (Bertsekas, 1987) that $\rho^\pi(x) = \rho^\pi(x')$ for all $x, x' \in \mathcal{X}$ under the assumption that the Markov chain for policy $\pi$ is ergodic. Here, we assume that the MDPs are ergodic for all stationary policies. Following the Q-notation of Watkins (1989), the optimal policy $\pi^*$ can be characterized in terms of the optimal average reward $\rho$ and the optimal relative Q-values $Q(x, a)$ for each state-action pair $(x, a)$ that are solutions to the following system of non-linear equations (Singh, 1994): for any state $x \in \mathcal{X}$ and action $a \in \mathcal{A}$,

$$Q(x, a) + \rho = r(x, a) + \sum_{y \in \mathcal{X}} \mathbf{Pr}(x'|x, a)\left[\max_{a'} Q(x', a')\right]$$
$$= r(x, a) + \mathbf{E}_{x'}\left[\max_{a'} Q(x', a')\Big|x, a\right]. \quad (4)$$

For this setup, the optimal policy is defined as $\pi^*(x) := \arg\max_a Q(x, a)$ for any state $x \in \mathcal{X}$.

## 3. Derivation of the Controller

One can exploit the Bellman optimality equations in (4) to define a space of optimal controllers. In particular, any $\rho \in \mathbb{R}$ and collection of Q-values $Q(x, a) \in \mathbb{R}$ where $x \in \mathbb{N}$ and $a \in \mathcal{A}$ characterize an optimal controller. Hence, one can parameterize the space of controllers with a vector $\theta \in \Theta := \mathbb{R}^\infty$ containing the average reward and all the Q-values. To apply the Bayesian control rule, we need to derive probabilistic models for actions and observations.

Noting that in cycle $t$ the controller issues an action $a_t \in \mathcal{A}$ and receives a reward $r_t \in \mathcal{R}$ and a state $x_t \in \mathcal{X}$, one can define the space of actions and observations for the Bayesian control rule as $\mathcal{A}$ and $\mathcal{O} := \mathcal{R} \times \mathcal{X}$ respectively.

Let $x = x_{t-1}$, $a = a_t$, $r = r_t$ and $x' = x_t$. Given the controller's parameter vector $\theta$, the only additional information needed to apply the optimal policy is given by the last state $x$. Hence, we impose the independence property

$$P(a_t, o_t|\theta, a_{<t}, o_{<t}) = P(a, r, x'|\theta, x).$$

Furthermore, this can be decomposed as a product of three conditional probabilities:

$$P(a, r, x'|\theta, x) = P(a|\theta, x)P(x'|\theta, x, a)P(r|\theta, x, a, x'). \quad (5)$$

The first term, i.e. the probability of action $a$ given $\theta, x$ and $a$, is given by:

$$P(a|\theta, x) = P(a|\{Q(x, a')\}_{a' \in \mathcal{A}})$$
$$= \begin{cases} 1 & \text{if } a = \arg\max_{a'} Q(x, a') \\ 0 & \text{else,} \end{cases} \quad (6)$$

which is just the action taken by the optimal policy $\pi^*$ in state $x$.

For the second term in (5), i.e. the state transition probabilities given the past interactions, we observe that the average reward $\rho$ and the Q-values $Q(x,a)$ encoded in $\theta$ do not provide enough information to encode the transition probabilities. Thus we conclude that they are independent of the parameter, that is:

$$\text{for any } \theta, \theta' \in \Theta, \quad P(x'|\theta, x, a) = P(x'|\theta', x, a). \quad (7)$$

Finally we derive $P(r|\theta, x, a, x')$, i.e. the probabilities of rewards given the past interactions *and* the next state $x$. Note that the reward function $r(x,a)$ is *not* parameterized by $\theta$, thus we cannot know the exact value of $r$ from $(\theta, x, a, x')$.

Let $\xi(x, a, x')$ be the *mean instantaneous reward* defined by

$$\xi(x, a, x') := Q(x, a) + \rho - \max_{a'} Q(x', a'). \quad (8)$$

This quantity represents the mean of the instantaneous reward $r(x,a)$ as estimated indirectly using the pre- and post-action Q-values. Indeed, it is seen from Equation (4) that

$$r(x, a) = \xi(x, a, x') + \nu, \quad (9)$$

where

$$\nu := \max_{a'} Q(x', a') - \mathbf{E}[\max_{a'} Q(x', a')|x, a].$$

Here, the term $\nu$ is a deviation from $r(x,a)$ that can be interpreted as random observation noise. Assuming that $\nu$ can be reasonably approximated by a normal distribution $\mathcal{N}(0, 1/p)$ with precision $p$, then we can write down a likelihood model for the immediate reward $r$ using the Q-values and the average reward, i.e.

$$P(r|\theta, x, a, x') = \sqrt{\frac{p}{2\pi}} \exp\left\{-\frac{p}{2}(r - \xi(x, a, x'))^2\right\}. \quad (10)$$

This completes our model of the controller with parameter vector $\theta$.

To apply the Bayesian control rule over the controllers in $\Theta$, the intervened posterior distribution $P(\theta|\hat{a}_{\leq t}, o_{\leq t})$ defined in Equation (3) needs to be computed. Fortunately, due to the simplicity of the likelihood model, one can easily devise a conjugate prior distribution.

Inserting the likelihood into Equation (3), one obtains

$$P(\theta|\hat{a}_{\leq t}, o_{\leq t})$$
$$= \frac{P(x'|\theta, x, a)P(r|\theta, x, a, x')P(\theta|\hat{a}_{<t}, o_{<t})}{\int_{\tilde{\Theta}} P(x'|\theta', x, a)P(r|\theta', x, a, x')P(\theta'|\hat{a}_{<t}, o_{<t})\, d\theta'}$$
$$= \frac{P(r|\theta, x, a, x')P(\theta|\hat{a}_{<t}, o_{<t})}{\int_{\tilde{\Theta}} P(r|\theta', x, a, x')P(\theta'|\hat{a}_{<t}, o_{<t})\, d\theta'}, \quad (11)$$

where we have replaced the sum by an integration over $\tilde{\Theta}$, the finite-dimensional real space containing only the average reward and the Q-values of the observed states, and where we have simplified the term $P(x'|\theta, x, a)$ because it is constant for all $\theta' \in \tilde{\Theta}$.

By inspection of Equation (11), one sees that $\theta$ encodes a set of independent normal distributions over the immediate reward having means $\xi(x, a, x')$ indexed by triples $(x, a, x') \in \mathcal{X} \times \mathcal{A} \times \mathcal{X}$. In other words, given $(x, a, x')$, the rewards are drawn from a normal distribution with unknown mean $\xi(x, a, x')$ and known variance $\sigma^2$. The sufficient statistics are given by $n(x, a, x')$, the number of times that the transition $x \to x'$ under action $a$, and $\bar{r}(x, a, x')$, the mean of the rewards obtained in the same transition. The conjugate prior distribution is well known and given by a normal distribution with hyperparameters $\mu_0$ and $\lambda_0$:

$$P(\xi(x, a, x')) = \mathcal{N}(\mu_0, 1/\lambda_0)$$
$$= \sqrt{\frac{\lambda_0}{2\pi}} \exp\left\{-\frac{\lambda_0}{2}\big(\xi(x, a, x') - \mu_0\big)^2\right\}. \quad (12)$$

The posterior distribution is given by

$$P(\xi(x, a, x')|\hat{a}_{\leq t}, o_{\leq t}) = \mathcal{N}(\mu(x, a, x'), 1/\lambda(x, a, x'))$$

where the posterior hyperparameters are computed as

$$\mu(x, a, x') = \frac{\lambda_0 \mu_0 + p\, n(x, a, x')\, \bar{r}(x, a, x')}{\lambda_0 + p\, n(x, a, x')}$$
$$\lambda(x, a, x') = \lambda_0 + p\, n(x, a, x'). \quad (13)$$

Finally, the conjugate distribution of the parameter vector $\theta$ is simply the product

$$P(\theta|\hat{a}_{\leq t}, o_{\leq t}) = \prod_{x,a,x'} P(\xi(x, a, x')|\hat{a}_{\leq t}, o_{\leq t})$$
$$\propto \exp\left\{-\frac{1}{2}\sum_{x,a,x'} \lambda(x, a, x')\big(\xi(x, a, x') - \mu(x, a, x')\big)^2\right\} \quad (14)$$

because the $\xi(x, a, x')$ are independent but at the same time functions of $\theta$ (Equation 8). Thus, the BCR-MDP controller is fully specified by the actions probabilities in Equation (6), the likelihood models in Equations (7) and (10), and the prior distribution (12).

## 4. Inference and Acting

Inference can be carried out by sampling $\theta$ from the posterior distribution in Equation (14). The actions issued by BCR-MDP are by-products of the inference process. Here we derive an approximate Gibbs sampler for $\theta$. We introduce the following symbols: $\theta^{-\rho}$ and $\theta^{-Q(x,a)}$ stand for the parameter set removing $\rho$ and $Q(x,a)$ respectively; $\mu$ and $\lambda$ are matrices collecting the values of the posterior hyperparameters $\mu(x,a,x')$ and $\lambda(x,a,x')$ respectively; and $M(x) := \max_a Q(x,a)$ is a shorthand.

Substituting $\xi(x,a,x')$ in Equation (14) by its definition (Equation 8) and conditioning on the Q-values, we obtain the conditional distribution of $\rho$:

$$P(\rho|\theta^{-\rho},\mu,\lambda) = \mathcal{N}(\bar{\rho},1/S) \qquad (15)$$

where

$$\bar{\rho} = \frac{1}{S}\sum_{x,a,x'} \lambda(x,a,x')(\mu(x,a,x') - Q(x,a) + M(x')),$$

$$S = \sum_{x,a,x'} \lambda(x,a,x').$$

The conditional distribution over the Q-values is more difficult to obtain, because each $Q(x,a)$ enters the posterior distribution both linearly and non-linearly through $\mu$. However, if we fix $Q(x,a)$ within the max operations, which amounts to treating each $M(x)$ as a constant within a single Gibbs step, then the conditional distribution can be approximated by

$$P(Q(x,a)|\theta^{-Q(x,a)},\lambda,\mu) \approx \mathcal{N}\left(\bar{Q}(x,a), 1/S(x,a)\right) \qquad (16)$$

where

$$\bar{Q}(x,a) = \frac{1}{S(x,a)}\sum_{x'} \lambda(x,a,x')(\mu(x,a,x') - \rho + M(x')),$$

$$S(x,a) = \sum_{x'} \lambda(x,a,x').$$

We expect this approximation to hold because the resulting update rule constitutes a contraction operation that forms the basis of most stochastic approximation algorithms (Mahadevan, 1996). As a result, the Gibbs sampler draws all the values from normal distributions. In each cycle of the adaptive controller, one can carry out several Gibbs sweeps to obtain a sample of $\theta$ to improve the mixing of the Markov chain. However, our experimental results have shown that a *single Gibbs sweep per state transition* performs reasonably well.

Once a new parameter vector $\theta$ is drawn, BCR-MDP proceeds by taking the optimal action given by Equation (6). The resulting algorithm is listed in Algorithm 4. Note that only the $\mu$ and $\lambda$ entries of the transitions that have occurred need to be represented explicitly; similarly, only the Q-values of visited states need to be represented explicitly.

---

**Algorithm 1** BCR-MDP Gibbs sampler.

Initialize entries of $\theta$, $\lambda$ and $\mu$ to zero.
Set initial state to $x \leftarrow x_0$.
**for** $t = 1, 2, 3, \ldots$ **do**
  { Interaction }
  Set $a \leftarrow \arg\max_{a'} Q(x,a')$ and issue $a$.
  Obtain $o = (r, x')$ from plant.

  {Update hyperparameters}
  $\mu(x,a,x') \leftarrow \frac{\lambda(x,a,x')\mu(x,a,x') + p\,r}{\lambda(x,a,x') + p}$
  $\lambda(x,a,x') \leftarrow \lambda(x,a,x') + p$

  {Gibbs sweep}
  Sample $\rho$ using (15).
  **for all** $Q(y,b)$ of visited states **do**
    Sample $Q(y,b)$ using (16).
  **end for**
  Set $x \to x'$.
**end for**

---

## 5. Preliminary Empirical Results

We have tested BCR-MDP in two toy examples: a grid-world domain, and on a suite of randomly generated MDPs. To give an intuition of the achieved performance, the results are contrasted with those achieved by R-learning. We have used the R-learning variant presented in Singh (1994, Algorithm 3) together with the uncertainty exploration strategy (Mahadevan, 1996). The corresponding update equations are

$$Q(x,a) \leftarrow (1-\alpha)Q(x,a) + \alpha\big(r - \rho + \max_{a'} Q(x',a')\big)$$
$$\rho \leftarrow (1-\beta)\rho + \beta\big(r + \max_{a'} Q(x',a') - Q(x,a)\big), \qquad (17)$$

where $\alpha, \beta > 0$ are learning rates. The exploration strategy chooses with fixed probability $p_{\exp} > 0$ the action $a$ that maximizes $Q(x,a) + \frac{C}{F(x,a)}$, where $C$ is a constant, and $F(x,a)$ represents the number of times that action $a$ has been tried in state $x$. Thus, higher values of $C$ enforce increased exploration.

**Grid-world domain.** In Mahadevan (1996), a grid-world is described that is especially useful as a test bed for the analysis of RL algorithms. For our purposes, it is of particular interest because it is easy to design experiments containing *suboptimal limit-cycles*.

Figure 1, panel (a), illustrates the $7 \times 7$ grid-world. A controller has to learn a policy that leads it from any initial location to the goal state. At each step, the agent can move to any adjacent space (up, down, left or right). If the agent reaches the goal state then

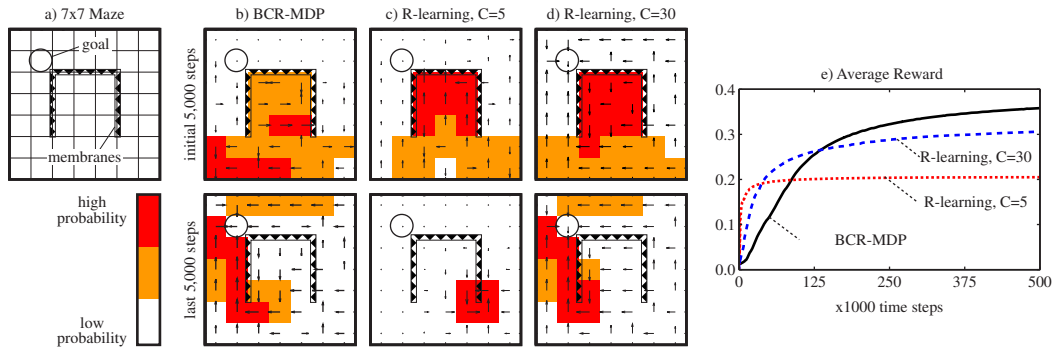

Figure 1. Results for the 7×7 grid-world domain. Panel (a) illustrates the setup. Columns (b)-(d) illustrate the behavioral statistics of the algorithms. The upper and lower row have been calculated over the first and last 5,000 time steps of randomly chosen runs. The probability of being in a state is color-encoded, and the arrows represent the most frequent actions taken by the agents. Panel (e) presents the curves obtained by averaging ten runs.

its next position is randomly set to any square of the grid (with uniform probability) to start another trial. There are also "one-way membranes" that allow the agent to move into one direction but not into the other. In these experiments, these membranes form "inverted cups" that the agent can enter from any side but can only leave through the bottom, playing the role of a local maximum. Transitions are stochastic: the agent moves to the correct square with probability $p = \frac{9}{10}$ and to any of the free adjacent spaces (uniform distribution) with probability $1 - p = \frac{1}{10}$. Rewards are assigned as follows. The default reward is $r = 0$. If the agent traverses a membrane it obtains a reward of $r = 1$. Reaching the goal state assigns $r = 2.5$.

The parameters chosen for this simulation were the following. For BCR-MDP, we have chosen hyperparameters $\mu_0 = 1$ and $\lambda_0 = 1$ and precision $p = 1$. For R-learning, we have chosen learning rates $\alpha = 0.5$ and $\beta = 0.001$, and the exploration constant has been set to $C = 5$ and to $C = 30$.

A total of 10 runs were carried out for each algorithm. The results are presented in Figure 1 and Table 1. R-learning only learns the optimal policy given sufficient exploration (panels c & d, bottom row), whereas BCR-MDP learns the policy successfully. In Figure 1e, the learning curve of R-learning is initially steeper than the Bayesian controller. However, the latter attains a higher average reward around time step 125,000 onwards. We attribute this shallow initial transient to the phase where the distribution over the operation modes is flat, which is also reflected by the initially random exploratory behavior.

To test wether the performance of BCR-MDP scales up with a larger problem, we have conducted a sec-ond grid-world experiment with where the number of states has roughly been doubled. The results for this 10×10 maze are illustrated in Figure 2. The reward for reaching the goal state has been set to $r = 10$ in this case. The precision for this experiment has been set to $p = 1/3$ to reflect higher uncertainty. This is still very low given that the range of possible rewards is $[0; 10]$. We have simulated one run of one million time steps. Again, one can see that the algorithm moves from a highly exploratory phase to an exploitative phase (Figure 2, left panels), eventually converging towards the optimal policy. The learning curve shows a steady increase in performance (Figure 2, right panel). Interestingly, around time step 300,000 the curve shows an abrupt change in slope. Presumably this is due to a change of the belief state: the algorithm was exploiting one of the two suboptimal limit cycles when it discovered the optimal limit cycle. This confirms our intuition, because the inference process cannot converge as long as there is still uncertainty over the policy space.

Table 1. Average reward attained by the different algorithms at the end of the run. The mean and the standard deviation has been calculated based on 10 runs.

|  | Average Reward |
|---|---|
| BCR-MDP | $0.3582 \pm 0.0038$ |
| R-learning, $C = 30$ | $0.3056 \pm 0.0063$ |
| R-learning, $C = 5$ | $0.2049 \pm 0.0012$ |

**Randomly generated MDPs.** The purpose of this experiment is to test the robustness of the algorithm under different environments. In this second test bed, random ergodic MDPs have been generated: a) with

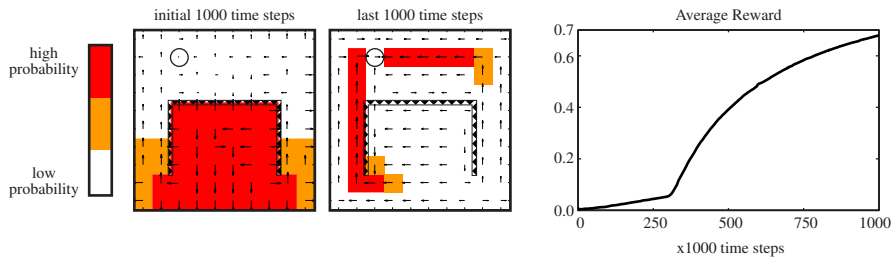

Figure 2. 10x10 maze task.

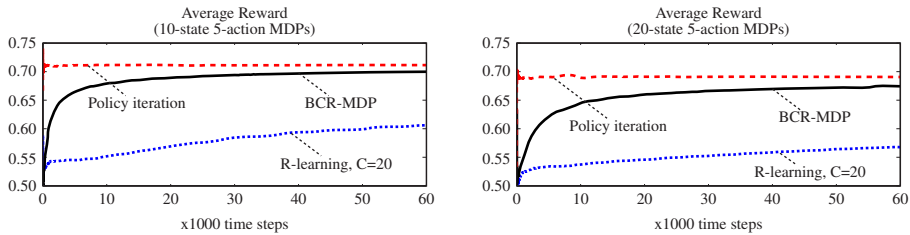

Figure 3. Comparison of the average reward for BCR-MDP, policy iteration and R-learning. A total of 60 different MDPs where randomly generated, one half having 10 states and 5 actions (left panel), and the other half having 20 states and 5 actions (right panel). The three algorithms where tested on these MDPs and their learning curves averaged.

Table 2. Average reward attained by the different algorithms at the end of the run. The mean and the standard deviation has been calculated based on 30 runs.

| Average Reward : | 10-state, 5-actions | 20-state, 5-actions |
|---|---|---|
| Policy iteration | $0.7114 \pm 0.0207$ | $0.6906 \pm 0.0101$ |
| BCR-MDP | $0.6998 \pm 0.0211$ | $0.6743 \pm 0.0102$ |
| R-learning, $C = 20$ | $0.6061 \pm 0.0216$ | $0.5677 \pm 0.0104$ |

10 states and 5 actions and b) with 20 states and 5 actions. The transition and payoff matrices have been constructed randomly: all transitions had non-zero probabilities and all rewards took on values in [0; 1]. In each run, a new MDP is generated and the three agents are tested on it: BCR-MDP with precision $p = 1$; R-learning with $\alpha = 0.5$, $\beta = 0.001$ and $C = 20$; and policy iteration. The latter has been used to estimate the maximum performance, i.e. the performance of an informed agent. We have simulated a total of 30 runs with 60,000 time steps for both cases and averaged the curves. The results, presented in Figure 3 and Table 2, show that BCR-MDP quickly approximates the optimal average reward, in both cases significantly faster than R-learning.

## 6. Summary and Conclusion

The reformulation of the adaptive control problem as the minimization of the relative entropy over the causal dependencies stated in Equation (1) leads to an explicit solution given by the Bayesian control rule. This rule constitutes a general method to construct adaptive controllers from plant-specific controllers. Its main advantage is that it allows replacing the intractable calculation of the optimal policy for the class of plants by an on-line inference procedure, where actions are simply by-products of the inference process. Conceptually, the Bayesian control rule instantiates several well-known ideas: the action selection strategy is a probability matching method (Wyatt, 1997); mixing task-optimal controllers is a mixture of experts technique (Jacobs et al., 1991); and minimizing the relative entropy to design a controller is equivalent to maximizing the compression of the controller's I/O stream (MacKay, 2003).

To illustrate the potential of the Bayesian control rule, we have derived BCR-MDP, an adaptive controller to solve undiscounted MDPs with finite state and action spaces and unknown dynamics. BCR-MDP is very simple to understand and to implement using the Gibbs sampler proposed in Section 4. Empirical results show that the built-in exploration-exploitation strategy avoids getting trapped in local minima.

Using Bayesian techniques in RL has a long history and has shown to be useful because they provide a systematic way of incorporating prior knowledge and domain assumptions into the problem and updating them as more data are observed. This allows quantifying the uncertainty of the quantity of interest, e.g. the value function, action-value function, etc. The idea of restating RL as an inference problem has also been proposed in Toussaint et al. (2006). This approach uses the expectation-maximization (EM) algorithm to infer the optimal policy, and special pruning techniques to reduce the computational complexity. It is interesting to point out that in the case of undiscounted MDPs, Bayesian Q-learning (Dearden et al., 1998) resembles closely BCR-MDP. Our contribution is to show that such an algorithm can be derived from a more general relative entropy minimization principle, including some features like the implicit exploration-exploitation trade-off.

We expect similar simplifications to hold for the design of adaptive controllers for other classes of plant dynamics. In particular, potential applications of the Bayesian control rule include extensions to continuous state and action spaces and to partially observable Markov processes.